\documentclass[sigconf]{acmart}

\usepackage{tcolorbox}
\usepackage{xcolor}
\usepackage{booktabs}
\usepackage{multirow}

\AtBeginDocument{%
  }

\setcopyright{acmlicensed}
\copyrightyear{2026}
\acmYear{2026}
\acmDOI{XXXXXXX.XXXXXXX}
\acmConference[Conference acronym 'XX]{Make sure to enter the correct
  conference title from your rights confirmation email}{June 03--05,
  2026}{Woodstock, NY}
\acmISBN{978-1-4503-XXXX-X/2018/06}

\begin{document}

\title{ThinkDeception: A Progressive Reinforcement Learning Framework for Interpretable Multimodal Deception Detection}

\author{Jinhao Song}
\affiliation{%
  \institution{Xi'an Jiaotong-Liverpool University}
  \city{Suzhou}
  \country{China}}
\email{2019201320@hrbeu.edu.cn}

\authornote{Both authors contributed equally to this research.}
\email{}
\author{Shan Liang}
\authornotemark[1]
\email{Shan.Liang@xjtlu.edu.cn}
\affiliation{%
  \institution{Xi'an Jiaotong-Liverpool University}
  \city{Suzhou}
  \country{China}
}

\author{Yiqun Yue}
\affiliation{%
  \institution{Xi'an Jiaotong-Liverpool University}
  \city{Suzhou}
  \country{China}}

\author{Zhuhuayang Zhang}
\affiliation{%
  \institution{Xi'an Jiaotong-Liverpool University}
  \city{Suzhou}
  \country{China}}

\author{Tianqi Gao}
\affiliation{%
  \institution{Xi'an Jiaotong-Liverpool University}
  \city{Suzhou}
  \country{China}}
\email{Tianqi.Gao25@student.xjtlu.edu.cn}

\renewcommand{\shortauthors}{Trovato et al.}

\begin{abstract}
Multimodal deception detection is critical for identifying fraudulent intentions, yet existing approaches predominantly rely on end-to-end black-box paradigms. These methods suffer from a severe lack of interpretability—failing to provide transparent reasoning trajectories and struggling to explicitly capture the subtle, cross-modal inconsistencies inherent in deceptive behaviors. To transcend these limitations, we propose ThinkDeception, a novel and interpretable multimodal deception detection framework. As a pioneering effort, it introduces Multimodal Large Language Models (MLLMs) into this domain, transforming deception detection from a traditional binary classification task into an explicit cognitive reasoning process. Facilitated by the first meticulously annotated step-by-step multimodal Chain-of-Thought (CoT) dataset, we develop a foundational model, ThinkDeception-Base, empirically validating the critical role of modal inconsistency in decoding deception. Building upon this foundation, our core innovation lies in proposing Visual-Audio Consistency Group Relative Policy Optimization(VAC-GRPO) equipped with a progressive training strategy. Distinct from standard GRPO, we stratify the training data into four progressive difficulty tiers, guiding the model through a psychologically grounded ``easy-to-hard'' cognitive transition. By innovatively coupling this dynamic curriculum scheduler with a multi-dimensional, process-aware reward mechanism and a reflective learning paradigm, we significantly elevate the model's overall reasoning quality. Extensive experiments on mainstream benchmarks demonstrate that ThinkDeception establishes a new state-of-the-art (SOTA), significantly outperforming existing methods in both detection accuracy and rationale quality. Ultimately, this work successfully drives the field of deception detection toward interpretable, multimodal cognitive reasoning.
\end{abstract}



\maketitle

\section{Introduction}
Although multimodal deception detection methods integrating visual and acoustic information have achieved significant progress, they remain constrained by two major bottlenecks\cite{Deception1}\cite{Deception2}. First, due to the high costs associated with data collection and annotation, existing datasets are limited in scale, rendering models prone to local overfitting and hindering their ability to extract universally applicable deceptive cues. Second, the diversity of data source scenarios(courtrooms,laboratory settings) introduces significant domain discrepancies\cite{MDPE}, coupled with the inherent heterogeneity across modalities, severely impedes the learning of multimodal representations and the cross-domain generalization capabilities of the models. With the rapid advancement of multimodal large language models(MLLMs) in areas such as video understanding\cite{mllms1}\cite{mllms2}\cite{mllms3}\cite{mllms4}\cite{5}\cite{6}\cite{7}, we also begin to contemplate a question: Can we fully leverage the reasoning potential of multimodal large language models(MLLMs) to propose a method that, akin to human cognition, progressively analyses the deceptive cues step by step and ultimately arrives at a judgement?

Therefore, we present the first exploration of an interpretable deception detection method empowered by Reinforcement Learning (RL). During our investigation, we identified and tackled the following core bottlenecks: (1) Lack of Fine-Grained Reasoning Datasets: Current datasets\cite{MDPE}\cite{dolos}\cite{RLTD}\cite{BOL} are largely limited to coarse-grained veracity labels or shallow feature shifts, severely lacking the fine-grained visual and acoustic descriptive annotations necessary to supervise the reasoning process. Most critically, the field still lacks high-quality datasets that can directly instruct RL models to engage in Chain-of-Thought (CoT) reasoning. (2) Inadequate Logical Reasoning Capabilities: Current MLLMs lack a systematic reasoning paradigm for deception detection\cite{yang2024emollm}\cite{fang2026emo}. They struggle to align critical multimodal cues, including visual micro-expressions and AUs intensities, acoustic pitch and prosody fluctuations, and textual emotional shifts. (3) Transfer Limitations of Traditional RL\cite{RL1}\cite{rl2}: While RL excels in visual understanding, its direct application to deception detection is still very limited. Relying strictly on final classification accuracy for outcome supervision creates a sparse reward environment. This frequently induces factual hallucinations in audio-visual cue extraction, severely compromising the interpretability and reliability of the reasoning chain. (4) Significant Heterogeneity and Domain Shifts: Existing datasets are highly heterogeneous and feature spontaneous, highly subtle deceptive behaviors. Applying RL directly to data with such severe domain shifts frequently traps models in local optima or precipitates training collapse.

\begin{figure}[h]
  \centering
  \includegraphics[width=\linewidth]{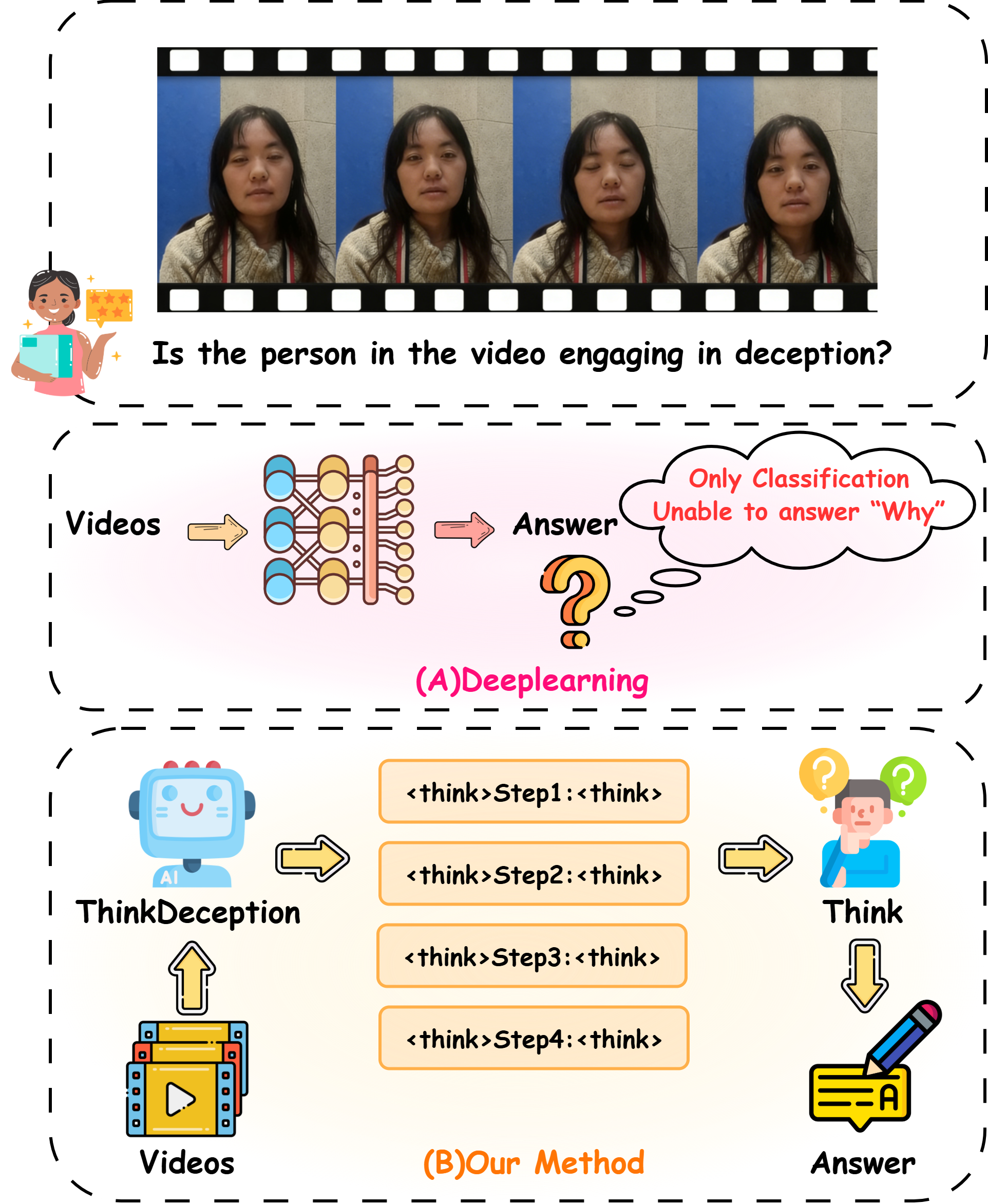}
  \caption{(A) Conventional deep learning directly yields predictions without interpretable processes.(B) Conversely, our method performs explicit step-by-step reasoning to deliver both transparent analytical trajectories and the final result.}
    \Description{}
\end{figure}

To address the aforementioned challenges, we propose \textbf{ThinkDeception}, a novel multimodal deception detection reasoning framework enhanced by Reinforcement Learning (RL). First, to bridge the critical gap in reasoning data, we construct \textbf{Deception-10K}, a high-quality multimodal Chain-of-Thought (CoT) dataset derived from open-source data. We design a specialized annotation pipeline to extract textual, visual, and acoustic cues, supplemented by fine-grained timestamp annotations to achieve precise audio-visual alignment. Second, as directly applying RL strategies to large models often suffers from convergence difficulties and fails to capture core cues, we adopt a ``teach-then-align'' paradigm. Using Supervised Fine-Tuning (SFT), we train a \textbf{ThinkDeception-Base} to initially acquire the capabilities of step-by-step reasoning and cross-modal inconsistency verification. This stage ensures that the model's reasoning process closely aligns with human psychological cognitive models. Finally, in the RL phase, we introduce a \textbf{Curriculum Learning}\cite{narvekar2020curriculum}\cite{zhao2026thinkdrive} mechanism. By categorizing the data into four difficulty levels—truthful, low-level, mid-level, and high-level deception—we guide the model learning through a progressive and easy to hard approach.  During this process, alongside foundational rule and format-based rewards, we pioneer a fine-grained, step-wise evaluation metric following the cognitive sequence of ``Observation-Listening-Reasoning-Answering''. By providing comprehensive reward supervision across the multimodal reasoning chain, we ultimately ensure the dual superiority of both reasoning quality and prediction accuracy.

\section{Related Work}

\subsection{Deception Detection}

Deception detection has recently achieved remarkable progress in single-modal representation and multimodal fusion. Specifically, capturing fine-grained acoustic features\cite{safe-qaq}, analyzing visual emotion and eye movement\cite{EmotionTF}\cite{eye}, and enhancing cross-domain generalization through feature alignment and unified mapping\cite{LCUNet} have significantly advanced the field.\cite{hu2024exploitingmultimodalspatialtemporalpatterns}

Nevertheless, current deep learning methods are hindered by their inherent black-box nature\cite{LCUNet}\cite{MMPDA}\cite{Cog}, producing predictions that lack traceable and verifiable logical reasoning grounds. Therefore, overcoming this interpretability bottleneck to shift the paradigm from opaque classification toward full-pipeline transparent reasoning is crucial for ensuring both the reliability and accuracy of deception detection.

\subsection{GRPO and Multimodal Reasoning}

Group Relative Policy Optimization (GRPO)\cite{DeepSeekMath} significantly reduces training overhead by estimating advantages via intra-group relative scores. DeepSeek-R1\cite{DEEPSEEK-R1} demonstrated that sparse rule-based rewards can elicit emergent Chain-of-Thought (CoT) reasoning. However, relying solely on outcome rewards makes standard GRPO prone to ``reward hacking'' in multimodal tasks, causing the model to generate superficially fluent reasoning disconnected from perceptual evidence, thereby undermining CoT credibility.

To address this, Vision-R1\cite{visionr1} and Video-R1\cite{videor1} extended GRPO to visual domains, demonstrating the generalization benefits of RL post-training. Subsequently, works like GRPO-CARE\cite{grpocare}, Fact-R1\cite{Fact-R1}, and EmotionThinker\cite{Emothinker} introduced process-aware rewards and progressive constraints, partially mitigating advantage signal collapse and providing a paradigm for multimodal reasoning.

While emerging works have successfully adapted RL-driven CoT to specific fields like autonomous driving (ThinkDrive\cite{zhao2026thinkdrive}) and emotion recognition (EmotionThinker\cite{Emothinker}, EMO-R3\cite{yang2024emollm}), they operate under the fundamental assumption of cooperative consistency across multimodal features. Deception detection, however, is an inherently adversarial cognitive process driven by deliberate behavioral camouflage.Therefore, the development of GRPO optimization mechanisms for deception detection is critical for advancing deceptive reasoning in multimodal large language models(MLLMs).

\section{Method}

\begin{figure*}
  \centering
  \includegraphics[width=\textwidth]{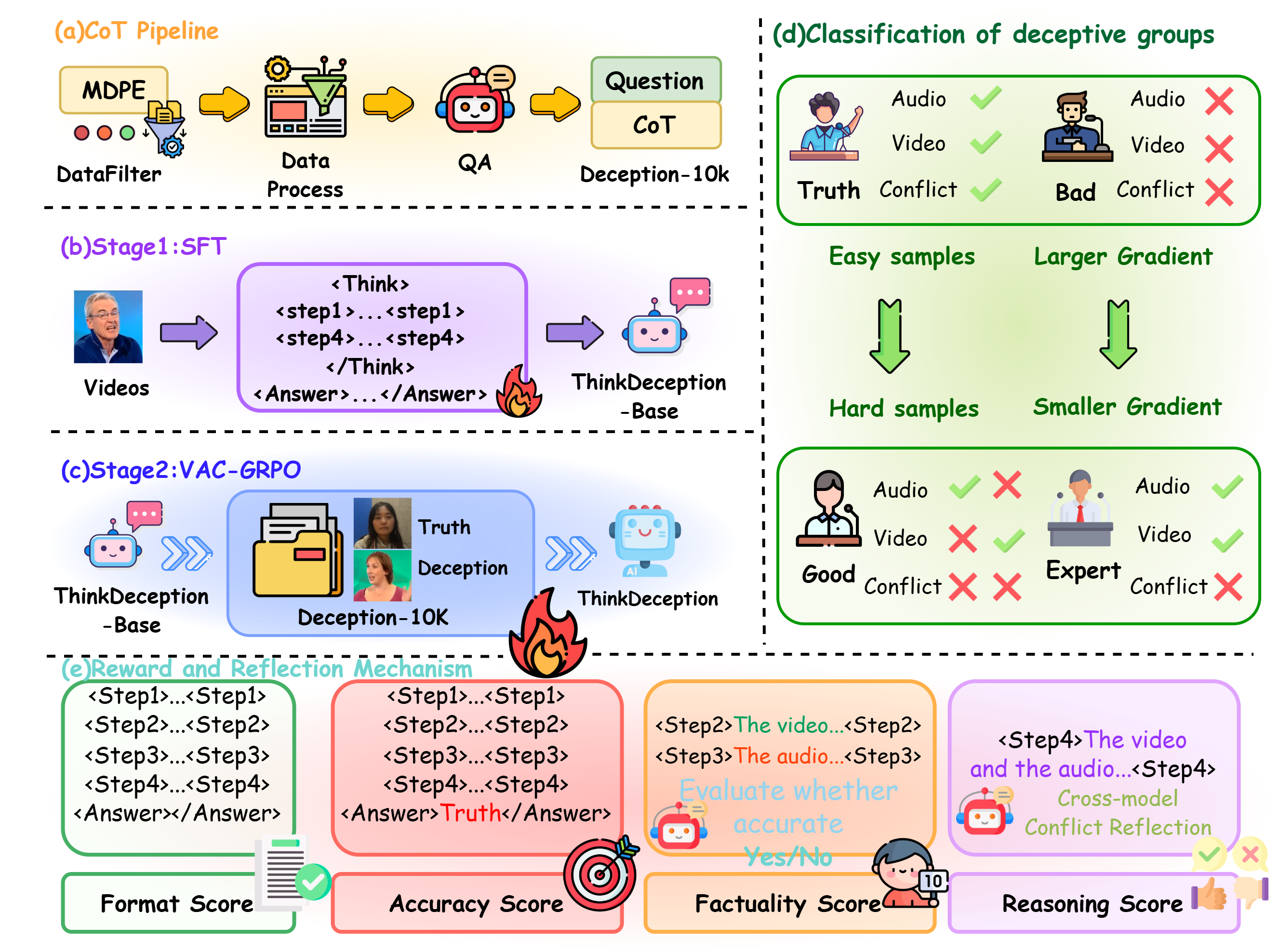}
  \caption{The overall pipeline of the proposed ThinkDeception framework. It comprises four main components: (a) Dataset Processing Pipeline; (b) Supervised Fine-Tuning (SFT) Phase; (c) Reinforcement Learning Phase; and (d) Progressive Training Strategy, which stratifies the dataset into four distinct difficulty levels to facilitate an easy-to-hard progressive curriculum; (e) Reward and Reflection mechanism}
  \Description{}
  \label{fig:Fig2}
\end{figure*}

The proposed ThinkDeception framework unfolds in three key stages, as depicted in Figure~\ref{fig:Fig2}. Initially, we formulate the Deception-10K dataset, which provides step-by-step reasoning trajectories grounded in fine-grained visual, acoustic, and temporal annotations. Subsequently, leveraging Qwen2.5-Omni-7B\cite{Qwen2.5-Omni} as the foundational architecture, we apply Supervised Fine-Tuning (SFT) for a cold start, producing ThinkDeception-Base equipped with initial deductive reasoning skills. Ultimately, we deploy a novel scheduled training strategy alongside a reflective Reinforcement Learning (RL) approach to comprehensively refine and optimize the model's reasoning optimization process.

\subsection{Deception-10K}

By combining open-source benchmarks including MDPE\cite{MDPE} and DOLOS\cite{dolos}, RLTD\cite{RLTD}, and Box of Lies\cite{BOL}, we construct the first fine-grained audio-visual Chain-of-Thought (CoT) dataset. This dataset comprises 10,000 video-reasoning pairs, totaling approximately 50 hours, with each sample featuring step-by-step reasoning trajectories and precise timestamp alignment annotations.

For visual feature extraction, we employ OpenFace3.0\cite{hu2025openface}, pre-trained on the Affect+ dataset\cite{AffectNet+}, to systematically extract facial Action Units (AUs) intensities and eight fundamental emotion categories. Instead of using emotion labels, we use the emotion probability distribution to represent the subtle and dynamic in this work. The core motivation behind this design is that both emotional expression and the leakage of deceptive cues are highly coherent temporal processes. Forcing the use of discrete labels inherently causes the model to overlook discriminative, latent emotional fluctuations. In contrast, continuous probability distributions effectively preserve these easily neglected yet crucial authentic emotional shifts.

In terms of acoustic features, we use standard speech processing tools to deeply disentangle and extract pitch, speech rate, and prosody directly from the raw audio signals. Ultimately, these fine-grained multimodal cues serve as conditional prompts for the Qwen3-Omni-30B\cite{Qwen3-Omni} model, driving it to generate high-quality, step-by-step reasoning processes. The generated reasoning chains are strictly standardized into \verb|<Think>|\verb|</Think>|, \verb|<Step>|\verb|</Step>| and \verb|<Answer>|\verb|</Answer>| structural formats. To mitigate potential inherent biases and factual hallucinations from the language models, all generated reasoning trajectories were rigorously reviewed and scored by professional psychologists. Comprehensive pipeline details and dataset exemplars are provided in the Appendix.

\subsection{Progressive Training Strategy}

While Supervised Fine-Tuning (SFT) successfully aligns the output format, the model's inherent reasoning capabilities remain suboptimal, necessitating Reinforcement Learning (RL)\cite{DeepSeekMath} for further policy optimization. However, standard group-relative RL algorithms struggle during early training. When confronted with complex, spontaneous deception, the model's initial reasoning deficits lead to incorrect trajectories, injecting severe gradient noise and risking catastrophic policy collapse. To overcome this bottleneck, we abandon traditional random sampling and introduce a progressive RL framework. By decoupling multimodal sample difficulty, we guide policy iteration through a psychologically grounded, ``easy-to-hard'' cognitive progression.

\subsubsection{\textbf{Multimodal Difficulty Assessment and Curriculum Sampling}}

As illustrated in Figure~\ref{fig:Fig2}(d), to facilitate a smooth cognitive transition during the Reinforcement Learning (RL)\cite{DeepSeekMath} phase, we design a difficulty assessment mechanism predicated on the salience of multimodal cues. Specifically, we categorize the samples into four progressive difficulty levels based on the deceptive features exhibited by the speakers: Truthful, Low-level deception, Mid-level deception, and High-level deception.

Let $y \in \{0, 1\}$ denote the ground-truth veracity label ($0$ for truthful, $1$ for deceptive). We define three boolean indicator variables, $I_v, I_a, I_c \in \{0, 1\}$, representing the presence of explicit deceptive cues in the visual modality, explicit deceptive cues in the acoustic modality, and significant cross-modal semantic-audio-visual conflicts, respectively. Consequently, for any given sample $x_i$, its difficulty level $d_i$ is formulated as follows:
\begin{equation}
d_i = 
\begin{cases} 
0, & \text{if } y = 0 \\
1, & \text{if } y = 1 \text{ and } (I_v = 1 \land I_a = 1) \\
2, & \text{if } y = 1 \text{ and } (I_v \oplus I_a = 1) \\
3, & \text{if } y = 1 \text{ and } (I_v = 0 \land I_a = 0 \land I_c = 1)
\end{cases}
\end{equation}

where $\oplus$ denotes the logical XOR operator. This formally models progressive deception concealment: low-level deceivers show explicit visual-acoustic flaws ($d_i=1$), mid-level deceivers reveal unimodal inconsistencies ($d_i=2$), and high-level deceivers fully camouflage audiovisual cues, requiring cross-modal inconsistency verification ($I_c=1$) to detect latent conflicts ($d_i=3$).

To prevent optimization instability caused by abrupt shifts in task complexity, we propose a Gaussian-weighted curriculum learning strategy. This smoothly prioritizes easier samples early on and dynamically transitions to harder ones. For $K=4$ difficulty levels, the unnormalized sampling weight $S_{Gaussian}(t,k)$ for difficulty $k \in \{0, 1, 2, 3\}$ at step $t$ is defined as:

\begin{equation}
S_{Gaussian}(t,k) = \exp\left(-\frac{(x_t - \mu_k)^2}{2\sigma^2}\right)
\end{equation}

The dynamic control variable $x_t$, which determines the temporal evolution of the sampling peak center, is calculated as:
\begin{equation}
x_t = \left(\frac{t}{T}\right)^\beta (K - 1)
\end{equation}
where $T$ represents the total number of RL training steps; $\mu_k = k$ denotes the fixed mean corresponding to difficulty level $k$; $\sigma$ is the variance parameter controlling the concentration of the sampling distribution; and $\beta$ is a non-linear modulation coefficient governing the drift rate of the sampling center $x_t$.

To obtain the actual batch sampling probability, we normalize the Gaussian weights across all difficulty levels. The final probability $P(k|t)$ of sampling a sample of difficulty $k$ at step $t$ is given by:
\begin{equation}
P(k|t) = \frac{S_{Gaussian}(t, k)}{\sum_{j=0}^{K-1} S_{Gaussian}(t, j)}
\end{equation}

By employing this progressive, difficulty-aware scheduling strategy, we effectively mitigate the issues of sparse rewards and training collapse caused by overly large exploration spaces in the initial RL stages. Concurrently, it compels the model to concentrate on resolving subtle cross-modal conflicts during the mid-to-late training phases, thereby profoundly activating the multimodal large language model's underlying reasoning potential.

\subsubsection{\textbf{Structured Analysis}}

Traditional deep learning for deception detection\cite{LCUNet}\cite{MMPDA}\cite{Cog} relies on implicit feature clustering, which suffers from overfitting and frequently misclassifies genuine stress as deceit. To overcome this, we propose a Structured Analysis and Reflection Mechanism that shifts to explicit step-wise reasoning. Because sophisticated deceivers meticulously camouflage their cues, deception rarely manifests in a single modality; rather, it is embedded in latent cross-modal inconsistencies. By forcing the model to scrutinize divergences across semantic, visual, and acoustic behaviors, we deconstruct the detection task into four reflection that mirror criminal psychology:

\begin{tcolorbox}[
    colback=blue!5!white,       
    colframe=blue!60!black,     
    boxrule=0.8pt, 
    arc=2pt,                    
    left=4pt, right=4pt, 
    top=4pt, bottom=4pt
]
\textcolor{blue!60!black}{\textbf{Structured Analysis and Reflection Mechanism:}}
\begin{itemize}
    \setlength{\itemsep}{3pt}
    \item \textbf{Textual Semantic Anchoring:} Extract and understand the textual content as the factual baseline and cognitive context of the entire video.
    \item \textbf{Visual Cue Decoding:} Focus on fine-grained visual changes within specific time segments, identifying potential deceptive cues such as masking smiles through precise description of micro-expressions and Action Units (AUs).
    \item \textbf{Acoustic Feature Mapping:} Analyze acoustic fluctuations such as pitch, speech rate, and prosody within corresponding time segments to quantitatively assess whether the speaker is in a state of abnormal tension or anxiety.
    \item \textbf{Cross-modal Conflict Reflection:} Act as the core reasoning hub to cross-reference and globally evaluate the aforementioned textual, visual, and acoustic outputs, aiming to unearth deep-seated cross-modal inconsistencies.
\end{itemize}
\end{tcolorbox}

\subsubsection{\textbf{Format Reward and Accuracy Reward}}

Specifically, following the rule-based reward paradigm of GRPO\cite{DeepSeekMath}, we define two reward terms to govern the model's output structure.

The format reward, denoted as $\mathcal{R}_{f}$, measures whether the model adheres to the structured output format. It verifies the presence of each intermediate reasoning step $s_i$ and ensures that the final answer is properly enclosed within the \verb|</Answer>| tags:
\begin{equation}
\mathcal{R}_{f} = 
\begin{cases} 
1, & \text{if the step and Answer format are correct;} \\ 
0, & \text{otherwise.} 
\end{cases}
\end{equation}

Meanwhile, the accuracy reward, denoted as $\mathcal{R}_{acc}$, evaluates whether the predicted deception label $\hat{\mathcal{E}}$ aligns with the ground-truth deception label $\mathcal{E}^*$:
\begin{equation}
\mathcal{R}_{acc} = 
\begin{cases} 
1, & \text{if } \hat{\mathcal{E}} = \mathcal{E}^*; \\ 
0, & \text{otherwise.} 
\end{cases}
\end{equation}

The two reward terms are jointly used to ensure that the generated reasoning contents can strictly meet the structural requirements.

\subsubsection{\textbf{Visual-Audio Consistency Reasoning Reward and Reflection Mechanism}}

The previously discussed reward mechanisms rely exclusively on outcome supervision, rendering the model highly susceptible to ``shortcut learning''. In such scenarios, the model may hallucinate flawed or irrational reasoning trajectories merely to manipulate the correct final answer. To overcome this limitation, we propose the \textbf{V}isual-\textbf{A}udio \textbf{C}onsistency Reward (VAC-GRPO), designed to impose fine-grained constraints on the factual accuracy and feature completeness of the audio-visual reasoning process. Specifically,  we leverage a knowledge distillation strategy to pre-train a lightweight judge model based on the Qwen2.5-Omni-3B\cite{Qwen2.5-Omni} architecture. We define a structured factual ground-truth set $J = \{F_v, F_a\}$ derived from raw videos, where $F_v$ contains continuous emotion probability distributions and AUs intensities, and $F_a$ includes disentangled pitch, speech rate, and prosody. To build the training corpus for this judge model, we prompt GPT-4o\cite{gpt} with the factual baseline $J$ to generate training data scored along two distinct dimensions: Factual Accuracy and Feature Completeness. To mitigate potential inherent biases and factual hallucinations from the language models, all generated reasoning trajectories were rigorously reviewed and scored by professional psychologists.

During the RL phase, the generated visual ($s_2$) and acoustic ($s_3$) reasoning steps are fed into the frozen judge model along with the factual baseline $J$ and specifically tailored evaluation prompts $P_v$ and $P_a$:

\begin{tcolorbox}[
    colback=green!5!white,     
    colframe=green!50!black,    
    boxrule=0.8pt,              
    arc=2pt,                   
    left=4pt, right=4pt,        
    top=4pt, bottom=4pt         
]
\textbf{Prompt $P_v$ (Visual Evaluation):} \\
Evaluate whether the following textual description accurately and comprehensively captures the visual content of the video in terms of factual alignment and feature completeness.
\end{tcolorbox}

\begin{tcolorbox}[
    colback=green!5!white,     
    colframe=green!50!black,    
    boxrule=0.8pt,              
    arc=2pt,                    
    left=4pt, right=4pt,       
    top=4pt, bottom=4pt         
]
\textbf{Prompt $P_a$ (Acoustic Evaluation):} \\
Evaluate whether the following textual description accurately and comprehensively captures the acoustic content of the video in terms of factual alignment and feature completeness.
\end{tcolorbox}

Conditioned on the factual ground-truth set $J$, the frozen judge model $\mathcal{M}_{judge}$ generates discrete reflective outputs for the visual and acoustic reasoning steps, respectively:
\begin{equation}
\hat{y}_{v} = \mathcal{M}_{judge}(F_v, s_2, P_v), \quad \hat{y}_{a} = \mathcal{M}_{judge}(F_a, s_3, P_a)
\end{equation}

The reflective outputs $\hat{y}_{v}$ and $\hat{y}_{a}$ are discrete binary indicators (``Yes'' or ``No''). Consequently, the modality-specific consistency rewards (denoted as $\mathcal{R}_{v}$ and $\mathcal{R}_{a}$) are formally defined as follows:
\begin{equation}
\mathcal{R}_{m} = 
\begin{cases} 
1, & \text{if } \hat{y}_{m} = \text{Yes;} \\
0, & \text{if } \hat{y}_{m} = \text{No.} 
\end{cases} \quad \text{for } m \in \{v, a\}
\end{equation}

By incorporating this dual-modality reward mechanism, we ensure that the generated reasoning process achieves superior quality across two critical dimensions: 
\begin{itemize}
    \item \textbf{Factuality:} It rigorously verifies whether the textual reasoning strictly adheres to the objective physical features delineated in $J$, such as subtle micro-expression variations and dynamic pitch shifts.
    \item \textbf{Completeness:} It assesses whether the model comprehensively captures and articulates the salient physical traits present in $J$.
\end{itemize}
Ultimately, this constraint guarantees a high degree of multimodal consistency between the step-wise textual descriptions and the actual audio-visual content.

However, for certain high-level deceivers, easily perceptible deceptive cues may not manifest within the audio-visual modalities. To address this, we propose a reflection reward mechanism based on cross-modal inconsistency verification.

Let $y \in \{0, 1\}$ denote the ground-truth label of the sample (where $1$ indicates the presence of deceptive behavior). Let $E \in \{0, 1\}$ be a boolean indicator variable representing whether the model extracts salient unimodal abnormal features during the generation of steps $s_1$ to $s_3$ ($E=1$ indicates the presence of explicit features). Furthermore, we define an indicator function $\Phi_{conflict}(s_4) \in \{0, 1\}$ to determine whether the model explicitly conducts logical reasoning and transition analysis regarding ``cross-modal conflicts, contradictions, or camouflage'' in the output $s_4$ of the fourth stage.

Based on this, the conditional logic alignment reward $R_{logic}$ is mathematically defined as follows:
\begin{equation}
R_{logic} = 
\begin{cases} 
+1.0, & \text{if } (y=1 \land E=0) \text{ and } \Phi_{conflict}(s_4)=1 \\
-1.0, & \text{if } (y=1 \land E=0) \text{ and } \Phi_{conflict}(s_4)=0 \\
+1.0, & \text{if } (y=1 \land E=1) \text{ and } \Phi_{conflict}(s_4)=1 \\
-1.0, & \text{if } y=0 \text{ and } \Phi_{conflict}(s_4)=1 \\
0.0,  & \text{otherwise}
\end{cases}
\end{equation}

This reward function effectively ensures that if no deceptive cues are captured during the initial three reasoning steps, the model is compelled to engage in deep reflective reasoning; conversely, if explicit deceptive cues are detected, it directly proceeds to output the conclusion. More detailed explanations are provided in the Appendix.

Ultimately, the joint reasoning reward is obtained by averaging the audio-visual consistency reward and the logic reflection reward:
\begin{equation}
R_{reasoning} = \frac{R_{a} + R_{v} + R_{logic}}{3}
\end{equation}

\begin{figure*}
  \centering
  \includegraphics[width=0.95\textwidth]{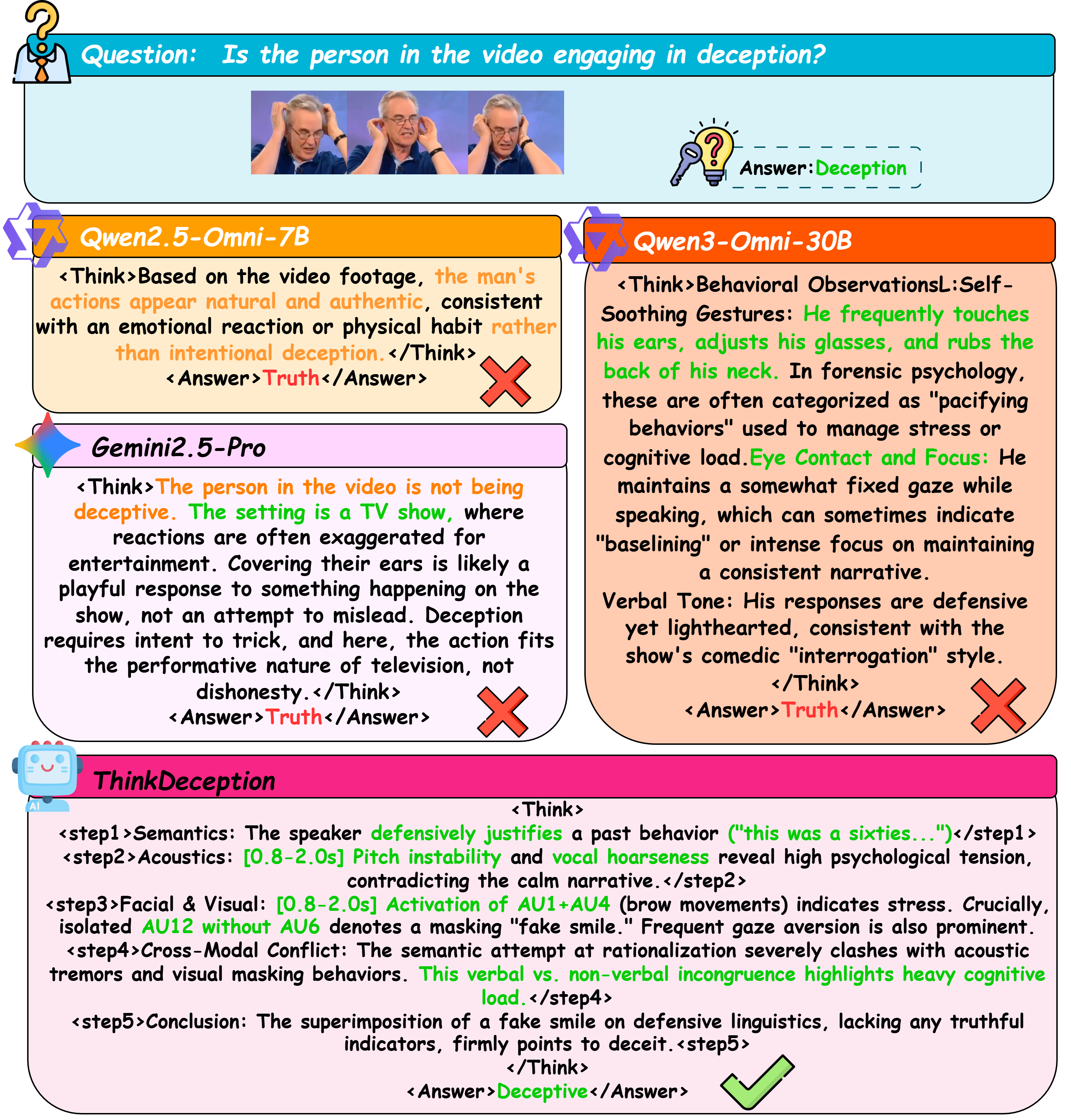}
  \caption{Qualitative comparison between ThinkDeception and baseline models.}
  \Description{}
  \label{fig:Over}
\end{figure*}

\begin{table*}[t]
    \centering
    \caption{Performance comparison of ThinkDeception with state-of-the-art baselines on four deception detection datasets. We report Classification Accuracy (\%) across both in-domain (DOLOS, MDPE) and cross-domain (RLTD, Box of Lies) settings, along with the Average Reasoning Quality Score (Avg RS, scaled 1-5).}
    \label{tab:main_results}
    \resizebox{\textwidth}{!}{
    \begin{tabular}{l ccc ccc c}
        \toprule
        & \multicolumn{2}{c}{\textbf{In-domain (ACC \%)}} & \multicolumn{2}{c}{\textbf{Cross-domain (ACC \%)}} & & \\
         \multirow{-2}{*}{\textbf{Methods}} & \textbf{DOLOS} & \textbf{MDPE} & \textbf{RLTD} & \textbf{Box of Lies} & \multirow{-2}{*}{\textbf{Avg ACC}} & \multirow{-2}{*}{\textbf{Avg RS}} \\
        \midrule
        
        \multicolumn{7}{l}{\textbf{\textit{Deep Learning Methods}}} \\
        \midrule
        LCUNet\cite{LCUNet} & \underline{70.56} & \underline{68.44} & 59.27 & 48.20 & 61.62 & - \\
        MMPDA\cite{MMPDA} & 68.95 & 66.45 & 60.33 & 56.92 & \underline{65.24} & - \\
        CogGuided\cite{Cog} & 69.18 & 65.91 & \underline{60.44} & \underline{58.75} & 65.18 & - \\
        \midrule
         \multicolumn{7}{l}{\textbf{\textit{Omni Large Language Models}}} \\
        \midrule
        Qwen2.5-Omni-7B\cite{Qwen2.5-Omni}  & 54.29 & 52.43 & 40.22 & 49.72 & 49.17 & 2.57 \\
        Qwen3-Omni-30B\cite{Qwen3-Omni}  & 53.33 & 54.21 & 38.13 & 50.26 & 48.98 & 2.58 \\
        Gemini2.5-Pro\cite{comanici2025gemini}  & 51.69 & 56.48 & 39.50 & 41.15 & 47.21 & 2.19 \\
        GLM-4.6v\cite{hong2025glm}  & 60.21 & 57.14 & 48.41 & 47.39 & 53.29 & 2.05 \\
        \midrule
        
        \multicolumn{7}{l}{\textit{Ours}} \\
        \midrule
        ThinkDeception-Base (SFT Only) & 60.15 & 59.43 & 55.62 & 56.80 & 58.00 & \underline{2.85} \\
        \textbf{ThinkDeception (Ours)} & \textbf{76.83} & \textbf{77.60} & \textbf{71.20} & \textbf{69.41} & \textbf{73.76} & \textbf{3.65} \\
        \bottomrule
    \end{tabular}
    }
\end{table*}

Ultimately, the overall optimization objective for the proposed VAC-GRPO framework is formulated as a weighted sum of the aforementioned reward components:
\begin{equation}
\mathcal{R}_{total} = \alpha_f \mathcal{R}_{f} + \alpha_{a} \mathcal{R}_{acc} + \alpha_{r} \mathcal{R}_{reasoning}
\end{equation}
where the coefficients $\alpha_f$, $\alpha_{a}$, and $\alpha_{r}$ represent the corresponding hyperparameter weights that govern the relative contribution of each reward component during the policy update.

\section{Experiments}

\subsection{Datasets and Evaluation Metrics}

To comprehensively evaluate the model's accuracy and reasoning capabilities in deception detection, our experiments are conducted on our newly constructed multimodal Chain-of-Thought dataset, Deception-10K. Specifically, we select four mainstream deception detection subsets encompassed by this dataset. Regarding the training protocol, we conduct independent model training on the DOLOS\cite{dolos} and MDPE\cite{MDPE} datasets (In-domain Training). Conversely, the remaining two datasets (RLTD\cite{RLTD} and Box of Lies\cite{BOL}) are strictly reserved as unseen test beds, dedicated exclusively to assessing the model's cross-domain generalization performance across diverse scenarios, racial demographics, and elicitation environments.

In terms of evaluation metrics, we adopt standard classification Accuracy (ACC) as the core quantitative metric for the final deception recognition results. Furthermore, as this paper pioneers the introduction of an explicit reasoning paradigm to this field, we incorporate reasoning quality metrics. By employing both an LLM-as-a-Judge and human expert blind reviews, we conduct a fine-grained quantitative assessment of the generated Chain-of-Thought (CoT) across critical dimensions, including factual consistency and logical coherence.

\subsection{Baseline Methods}

To thoroughly demonstrate the efficacy of the ThinkDeception framework, we compare it against 7 representative baseline methods, categorized into two groups:
\begin{itemize}
    \item Traditional Multimodal Deep Learning Methods: We select LCUNet\cite{LCUNet}, MMPDA\cite{MMPDA}, and CogGuided\cite{Cog}. To maintain absolute experimental fairness, we strictly isolate the Chain-of-Thought (CoT) texts in Deception-10K during the training of these models. Consequently, they are trained solely on the raw audio-visual features coupled with binary veracity labels.
    \item   Omni Large Language Models: Qwen2.5-Omni-7B\cite{Qwen2.5-Omni}, Qwen3-Omni-30B\cite{Qwen3-Omni}, Gemini2.5-Pro\cite{comanici2025gemini}, and GLM-4.6v\cite{hong2025glm}.
\end{itemize}

\subsection{Implementation Details}

All training procedures are conducted on $ 8\times$ NVIDIA A100 (80GB) GPUs. In the supervised fine-tuning (SFT) cold-start phase, we adopt Qwen2.5-Omni-7B\cite{Qwen2.5-Omni} as the foundation model and fine-tune it for one epoch on a subset of the Deception-10K dataset. This yields the baseline model, ThinkDeception-Base, which is equipped with preliminary step-by-step reasoning capabilities. During the subsequent reinforcement learning (RL) phase, we employ the Group Relative Policy Optimization (GRPO) algorithm\cite{DEEPSEEK-R1}\cite{ramesh2024group} with a learning rate of $1 \times 10^{-6}$. For each input video-text pair, the policy model generates $K=8$ candidate reasoning trajectories (rollouts), with the sampling process executed every 50 training steps.

\subsection{Evaluation Metrics}

To comprehensively and rigorously assess the generalization capability of our model, we conduct evaluations under both in-domain and cross-domain settings. Across all benchmark datasets, we employ classification accuracy as the primary objective metric. Furthermore, specifically tailored for Multimodal Large Language Models (MLLMs), we introduce a novel Reasoning Quality Score to quantitatively measure the logical coherence and cross-modal factual consistency of the generated rationales.

\subsection{Comparative Results}

As illustrated in Table~\ref{tab:ablation_study}, ThinkDeception achieves state-of-the-art (SOTA) performance in both overall detection accuracy and reasoning quality. Specifically, our model consistently secures the highest accuracy across all evaluated datasets, reaching an average accuracy of 73.76\% and outperforming the second-best baseline by a substantial absolute margin of 8.52\%.Notably, while the majority of existing baseline models excel in general visual understanding tasks, their performance in deception detection hovers around the random guess baseline of 50\%, despite being guided by identical prompts. This profound degradation underscores the significant potential and critical necessity for developing domain-specific Multimodal Large Language Models (MLLMs) tailored for deception detection. Furthermore, our model demonstrates exceptional robustness in cross-domain evaluations, particularly on the highly challenging, multi-speaker Box of Lies (BOL)\cite{BOL} dataset. This compellingly verifies that rather than merely overfitting to surface-level feature mappings, ThinkDeception has successfully internalized a generalized and unified reasoning paradigm for deception recognition.

\begin{figure}[h]
  \centering
  \includegraphics[width=\linewidth]{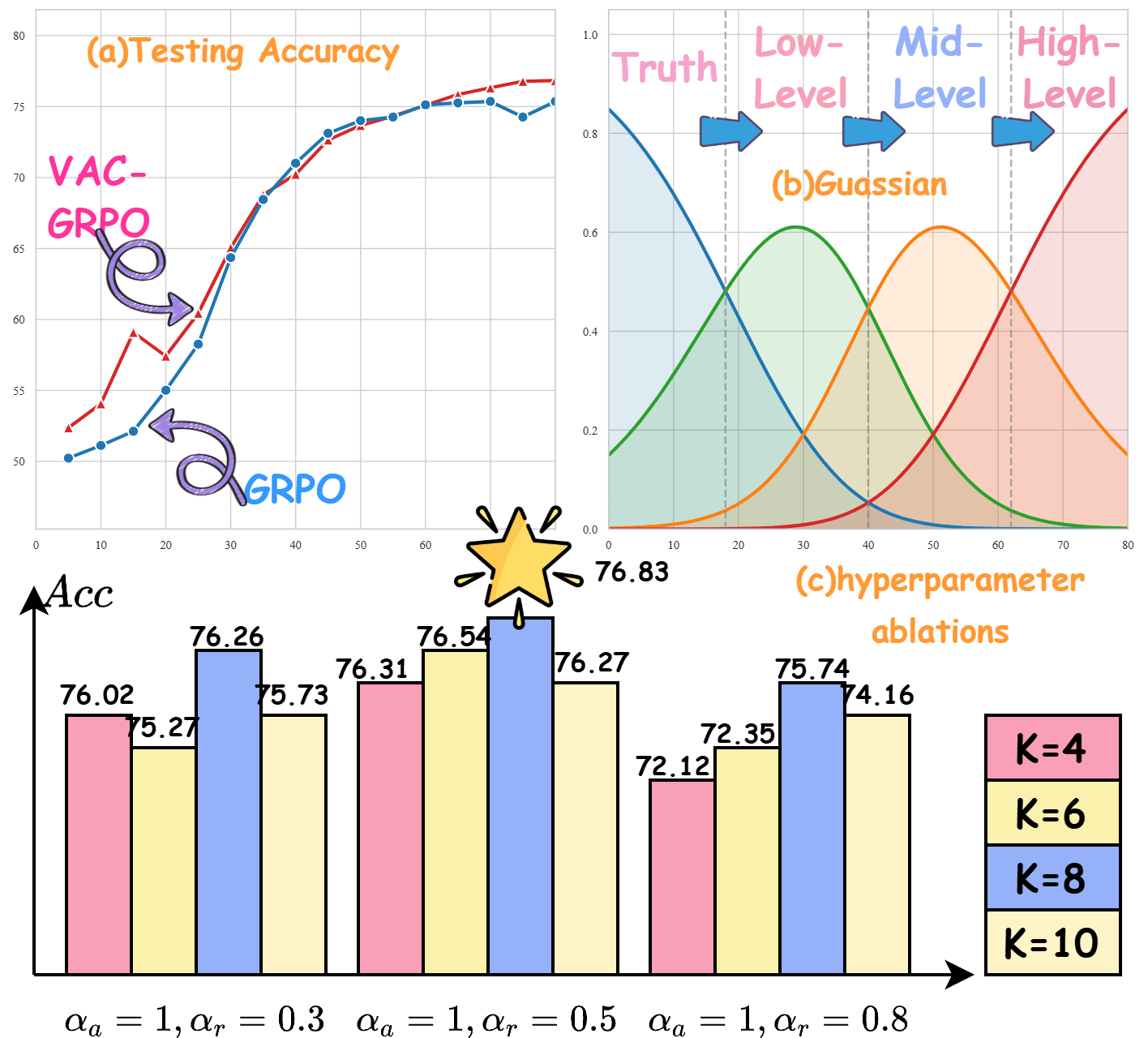}
  \caption{(a) Test accuracy comparison between VAC-GRPO and standard GRPO. (b) Dynamic sampling distribution across distinct difficulty levels over training steps. (c) Ablation study on key hyperparameters.}
    \Description{}
    \label{over:fig4}
\end{figure}

\subsection{Ablation Study}

Experimental results demonstrate that Supervised Fine-Tuning (SFT) yields a notable improvement in accuracy, substantiating the efficacy of our proposed modality inconsistency verification. The subsequent integration of VAC-GRPO reinforcement learning further elevates the model's performance. Furthermore, ablation studies on the core reward mechanisms reveal a phenomenon highly consistent with deceptive psychology: low-level visual-audio conflicts are inherently more discriminative than pure textual logic. This empirical finding corroborates that while deceivers can often fabricate logically watertight lies, it is exceedingly difficult for them to simultaneously suppress the physiological tension manifested in their visual and acoustic cues. This aligns with earlier observations that over-reliance on textual priors in deception detection makes the model highly susceptible to overfitting.

\begin{table}[t]
    \centering
    \caption{Ablation studies on the proposed ThinkDeception framework.}
    \label{tab:ablation_study}
    \resizebox{\linewidth}{!}{
    \begin{tabular}{l ccc ccc c}
        \toprule
        \multirow{2}{*}{\textbf{Model Variants}} & \multicolumn{2}{c}{\textbf{In-domain}} & \multicolumn{2}{c}{\textbf{Cross-domain}} & \multirow{2}{*}{\textbf{Avg ACC}} & \multirow{2}{*}{\textbf{Avg RS}} \\
        \cmidrule(lr){2-3} \cmidrule(lr){4-5}
        & \textbf{DOLOS} & \textbf{MDPE} & \textbf{RLTD} & \textbf{Box of Lies} & & \\
        \midrule
        
        \multicolumn{7}{l}{\textbf{\textit{(a) Ablation on Training Strategies}}} \\
        \midrule
        Qwen2.5-Omni-7B & 54.29 & 52.43 & 40.22 & 49.72 & 49.17 & 2.57 \\
        ThinkDeception-Base (SFT) & 60.15 & 59.43 & 55.62 & 56.80 & 58.00 & 2.85 \\
        ~~+ Standard GRPO ($\mathcal{R}_{f} + \mathcal{R}_{acc}$) & 71.30 & 72.52 & 63.52 & 62.31 & 67.41 & 3.27 \\
        ~~+ Progressive RL (\textbf{Full Model}) & \textbf{76.83} & \textbf{77.60} & \textbf{71.20} & \textbf{69.41} & \textbf{73.76} & \textbf{3.65} \\
        \midrule
        
        \multicolumn{7}{l}{\textbf{\textit{(b) Ablation on Reward Components}}} \\
        \midrule
        \textbf{ThinkDeception (Full Model)} & \textbf{76.83} & \textbf{77.60} & \textbf{71.20} & \textbf{69.41} & \textbf{73.76} & \textbf{3.65} \\
        ~~$w/o$ Visual-Audio Consist. ($\mathcal{R}_{av}$) & 70.13 & 69.67 & 67.82 & 69.40 & 69.26 & 3.15 \\
        ~~$w/o$ Logic Alignment ($\mathcal{R}_{logic}$) & 72.80 & 70.13 & 69.57 & 71.12 & 70.91 & 3.22 \\
        \bottomrule
    \end{tabular}
    }
\end{table}

Finally, we conduct hyperparameter ablations. Results indicate that the model achieves optimal performance when the number of sampled trajectories is set to $K=8$ . Additionally, sensitivity analysis on $\alpha_{a}$ and $\alpha_{r}$ shows that the peak performance occurs at $\alpha_{r}=0.5$. An excessively high $\alpha_{r}$ leads to performance degradation, demonstrating that overemphasizing intermediate reasoning signals can interfere with the advantage estimation of the primary task, thereby introducing optimization instability. This finding profoundly underscores the critical importance of maintaining a dynamic balance in reward distribution during multimodal reinforcement learning.

\subsection{Qualitative Analysis}

As illustrated in Figure~\ref{over:fig4}, compared to state-of-the-art models such as Qwen2.5-Omni-7B\cite{Qwen2.5-Omni}, Gemini 2.5 Pro\cite{comanici2025gemini}, and Qwen3-Omni-30B\cite{Qwen3-Omni}, our ThinkDeception framework achieves the highest results in both reasoning quality and factual consistency. The baseline models generally suffer from short-circuit reasoning, erroneous classifications, and severe hallucination issues where the reasoning trajectories are disconnected from factual evidence.

\subsection{Reliability of The Foundational Models}

Comprehensive details and reliability evaluations for the adopted generative (Qwen3-Omni-30B\cite{Qwen3-Omni}) and judge (GPT-4o\cite{gpt}, Qwen2.5-Omni-3B\cite{Qwen2.5-Omni}) models, as well as how to train a lightweight judge model\cite{judge}, are detailed in the Appendix.

\section{Conclusion}

In this paper, we propose ThinkDeception, successfully introducing the reasoning capabilities of Large Language Models into the domain of deception detection for the first time. This work drives a paradigm shift in deception recognition from a traditional binary classification task to an interpretable reasoning process. During the training phase, we design a progressive learning strategy to guide the model in internalizing deceptive features in an easy-to-hard manner. Concurrently, we introduce VAC-GRPO, enabling the model to conduct rigorous step-by-step reasoning. This architecture empowers the model to generate factually consistent reasoning steps and precisely capture latent deceptive cues. Comprehensive experiments across multiple benchmark datasets demonstrate that ThinkDeception establishes a new state-of-the-art in both detection accuracy and reasoning quality. 


\bibliographystyle{ACM-Reference-Format}
\bibliography{sample-base}


\end{document}